\documentclass[11pt,a4paper]{article}
\usepackage[hyperref]{acl2018}
\usepackage{times}
\usepackage{latexsym}
\usepackage{graphicx}
\usepackage{listings}
\usepackage{url}

\aclfinalcopy 

\title{Generating Clues for Gender based Occupation De-biasing in Text}

\author{Nishtha Madaan\textsuperscript{1}, Gautam Singh\textsuperscript{1}, Sameep Mehta\textsuperscript{1}, Aditya Chetan\textsuperscript{2}, Brihi Joshi \textsuperscript{2}\\ 
\textsuperscript{1}IBM Research-INDIA\\
\textsuperscript{2}IIIT-Delhi\\
\{nishthamadaan, gautamsi, sameepmehta\}@in.ibm.com\\ \{aditya16217, brihi16142\}@iiitd.ac.in}

\date{}

\begin{document}
\maketitle
\begin{abstract}

Vast availability of text data has enabled widespread training and use of AI systems that not only learn and predict attributes from the text but also generate text automatically. However, these AI models also learn gender, racial and ethnic biases present in the training data.\\
In this paper, we present the first system that discovers the possibility that a given text portrays a gender stereotype associated with an occupation. If the possibility exists, the system offers counter-evidences of opposite gender \emph{also} being associated with the same occupation in the context of user-provided geography and timespan. The system thus enables text de-biasing by assisting a human-in-the-loop. The system can not only act as a text pre-processor before training \emph{any} AI model but also help human story writers write stories free of occupation-level gender bias in the geographical and temporal context of their choice.



\end{abstract}

\section{Introduction}

AI systems are increasing and Natural Language Generation is getting ever more automated with emerging creative AI systems. These creative systems rely heavily on past available textual data. But often, as evident from studies done on Hollywood and Bollywood story plots and scripts, these texts are biased in terms of gender, race or ethnicity. Hence there is a need for a de-biasing system for textual stories that are used for training these creative systems.

Such de-biasing systems may be of two types 1) an end-to-end system that takes in a biased text and returns an unbiased version of it or 2) a system with a human-in-the-loop that takes a text, analyzes it and returns meaningful clues or pieces of evidence to the human who can appropriately modify the text to create an unbiased version. Since multiple types of biases may exist in the given text, the former de-biasing system requires identifying which biases to focus on and how to paraphrase or modify the sentence to de-bias it. These notions can often be subjective and it might be desirable to have a human-in-the-loop. This is the focus of the latter de-biasing system as well as the approach taken by us in the paper. 

Gender stereotyping with respect to occupations is one of the most pervasive biases that cuts across countries and age groups \cite{madaan2018analyze}. In this paper, we focus on de-biasing with respect to gender stereotyping in occupations.  This bias has also been recently noted in machine translation systems \cite{caliskan2016semantics}. In this translation tool, the sentences ``He is a nurse. She is a doctor" were translated from English to Turkish and back to English which inappropriately returned ``She is a nurse. He is a doctor"! 

In this paper, our system \footnote{https://youtu.be/Bvsd9WYd2U0} takes a piece of text and finds mentions of named entities and their corresponding occupations. From the gender of the named entities, the system suggests examples of real people with alternate gender who also had the corresponding occupation.

The rest of the paper is organized as follows - Section 2 describes the related work, Section 3 discusses about the design and Section 4 lays out the implementation of our de-biasing system. In Section 5 we describe a walk-through of our system and in Section 6 we conclude our paper.

\section{Past Work and Motivation}
\textbf{Analysis of gender bias in machine learning} in recent years has not only revealed the prevalence of such biases but also motivated much of the recent interest and work in de-biasing of ML models. \cite{zhao2017men} have pointed to the presence of gender bias in structured prediction from images. \cite{fast2016shirtless, madaan2018analyze} notice these biases in movies while \cite{gooden2001gender, millar2008selective} notice the same in children books and music lyrics.

\textbf{De-biasing the training algorithm} as a way to remove the biases focusses on training paradigms that would result in fair predictions by an ML model. In the Bayesian network setting, Kushner et al. have proposed a latent-variable based approach to ensure counter-factual fairness in ML predictions.  Another interesting technique (\cite{beutel2013copycatch} and \cite{zhang2016adversarial}) is to train a primary classifier while simultaneously trying to "deceive" an adversarial classifier that tries to predict gender from the predictions of the primary classifier.

\textbf{De-biasing the model after training} as a way to remove bias focuses on "fixing" the model after training is complete. \cite{bolukbasi2016man} in their famous work on gender bias in word embeddings take this approach to "fix" the embeddings after training.

\textbf{De-biasing the data at the source} fixes the data set before it is consumed for training. This is the approach we take in this paper by trying to de-bias the data or suggesting the possibility of de-biasing the data to a human-in-the-loop. A related task is to modify or paraphrase text data to obfuscate gender as in \cite{reddy2016obfuscating} Another closely related work is to change the style of the text to different levels of formality as in \cite{rao2018dear}.

\section{System Design}

\subsection{System Overview}

\begin{figure}
  \centering
  \includegraphics[width=\linewidth]{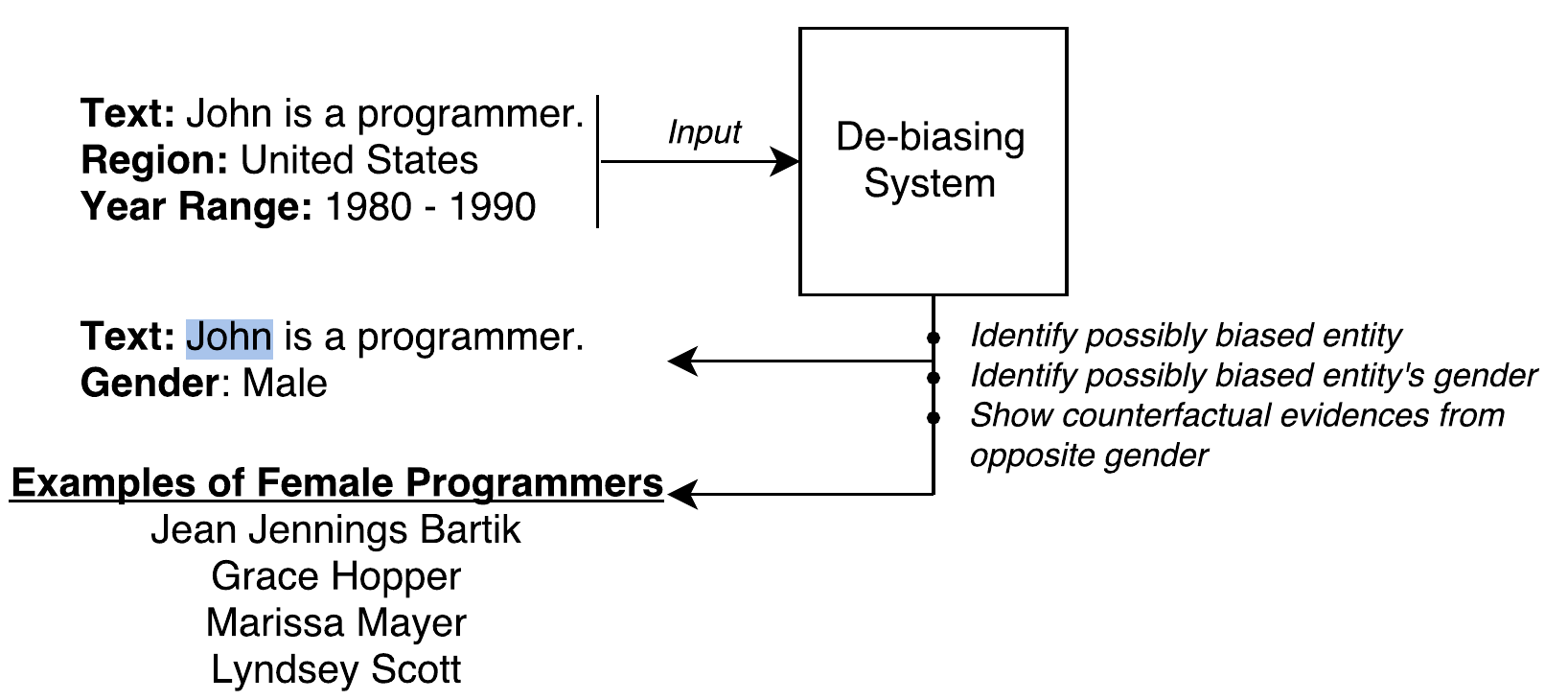}
  \caption{System Diagram for Text De-biasing}
  \label{fig:system_diagram}
\end{figure}

Our system allows the user to input a text snippet and choose the timespan and the demographic information. It highlights the named entities and their occupations which have a possibility of being biased. Further, the system outputs pieces of evidence in the form of examples of real people with that occupation from the selected time frame and region but having the opposite gender as shown in figure \ref{fig:system_diagram}

Our de-biasing algorithm is capable of tagging 996 occupations gathered from different sources*. A user who uses our de-biasing system can utilize the time-frame and region information to check for bias in a particular text snippet.  The detected bias can be shown to the user with pieces of evidence that can be then used to revisit the text and fix it. 

\section{System Implementation}

\subsection{Dataset Collection}

Our dataset comprises of the following - 1)  Occupation Data 2) Names Data. We will iterate over each of this one by one. 

\begin{enumerate}
    \item  Occupation Data: We gathered occupation lists from different sources on the internet including crowdsourced lists and government lists. Then, we classified the occupations into 2 categories - gender-specific occupation and gender-neutral occupations. These are used in the algorithm for bias checking which will be explained in the next sub-section. 

    \item Names Data:  We created a corpus of 5453 male and 6990 female names sourced from [ref: CMU repository of names]. For the dataset to map names to a gender, we referred to the NLTK data set and the records of baby names and their genders.
\end{enumerate}

\begin{figure*}[t]
  \centering
  \includegraphics[width=\linewidth]{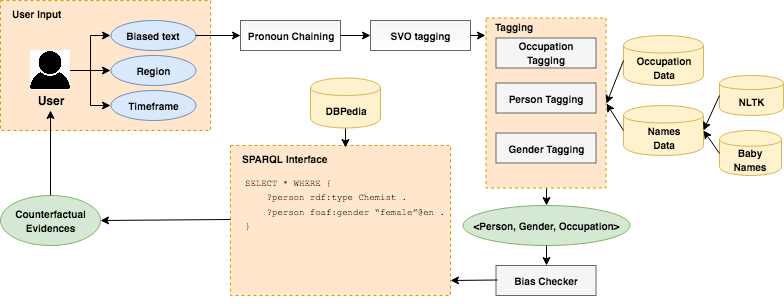}
  \caption{System Diagram - Occupation De-biasing System}
  \label{fig:sys-diag}
\end{figure*}

\subsection{Methodology}

Our system is represented in figure \ref{fig:sys-diag}. We have the following components in our system - 

\subsubsection{Pronoun Chaining} 
The task of mapping occupations to named entity or a person is crucial to perform debiasing on the text.  Often, the occupation of a person is mentioned with linking to the pronouns than the named entity itself. Hence, there is a need to resolve these co-references. We employ pronoun chaining using spaCy and replace the name of the pronoun with the named entity in the text entered by the user.

\subsubsection{SVO Tagging}
After we have done co-referencing, we parse the text to identify Subject, Verb, Object tuples. These tuples are further used to associate subjects i.e. named entity with its occupation i.e. object.

\subsubsection{Tagging}
We employ 3 specific types of tagging in our system - 
\begin{enumerate}
    
 \item \textbf{Occupation Tagging} - We use a dictionary based tagging mechanism to annotate occupation mentions in the text using the occupation dataset described in the previous section.

 \item \textbf{Person Tagging} - We use a dictionary based tagging for annotating person names in the text using the Names Dataset described in the previous section. 

 \item \textbf{Gender Tagging} - We further use the names dataset to resolve the genders of the persons identified in the previous person tagging step.
\end{enumerate}

At the end of this step, we obtain a set of 3-tuples $<$person, gender, occupation$>$.

\subsubsection{Bias Checker}
In this step, the goal is to check if $<$named entity, gender, occupation$>$ is potentially biased. This is done by first checking if the mentioned occupation is gender specific or gender neutral. If the occupation is gender specific, then we can clearly say it is free of bias. Otherwise, if the occupation is gender neutral, we try to fetch evidence examples of both genders performing that occupation in the given timeframe and demography. If we find no examples matching the query of the opposite gender, then we say that the text is free of bias.  Else, the system flags the sentence by highlighting the named entity and occupation and notifies the user about the possibility of bias.

\subsubsection{SPARQL Interface for fetching evidence }
In this section, we describe how we used SPARQL queries to fetch instances of people in DBpedia which belong to a certain gender, who lived in a certain time-frame and region and worked on a certain occupation.

In code-block below, we write a sample query that returns evidences of all female Chemists who were born in a city in US. The query returns 3-tuples containing the person's name, birth city, birth date and death date.

\begin{footnotesize}
\begin{lstlisting}[frame=single] 
SELECT * WHERE {
    ?person rdf:type "Chemist"@en 
    ?person foaf:gender "female"@en .
   
    ?person dbo:birthPlace ?bCity .
    ?bCity dbo:country "USA"@en .
  
    ?person dbo:birthDate ?bDate .
    ?person dbo:deathDate ?dDate .
}
\end{lstlisting}
\end{footnotesize}

As the next step, we filter these 3-tuple responses by checking if the life of the person (demarcated by the period between the birth and death dates) overlaps with the time-frame given by the user as input.

\section{Tool Walk-through using an example}

Consider a story-writer as a user of our system. The task is to be able to write bias free stories which are liked by viewers and earns high revenue in the BOX office. Here are few scenarios where this system can be used to identify bias.

\subsection{Scenario 1 : Year 1980-2000 in US}

The story-writer plans to write a story based in United States of America between timeframe 1980-2000. The story-writer uses our system and types in the natural language story -

\begin{footnotesize}
\begin{lstlisting} [frame= single, caption={Input of Scenario 1},label={lst:in-scene-1}]
John is a doctor. He treats his 
patients well. One day, he fell 
sick and started thinking about 
what he had been doing his whole 
life. 
\end{lstlisting}
\end{footnotesize}

This story interacts with our backend system and identifies if the story contains any occupational bias. Here, John is the named entity and doctor is the associated occupation. Furthermore, the system identifies John as a male character. It tries to search in backend if 'doctor' is a gender specific occupation or a gender neutral occupation. After detecting that it is a gender neutral occupation, the system checks the DBpedia corpus from 1980-2000 and fetches the instances of female doctors in the same timeframe in the United States. It displays the evidences for the user to go back and revisit and rewrite the story as below.

\begin{footnotesize}
\begin{lstlisting}[frame= single, caption={Modified version of story},label={lst:in-scene-2}] 
Mary is a doctor. She treats her 
patients well. One day, she fell 
sick and started thinking about 
what she had been doing her whole 
life.
\end{lstlisting}
\end{footnotesize}

\begin{figure}[t]
  \centering
  \includegraphics[width=\linewidth]{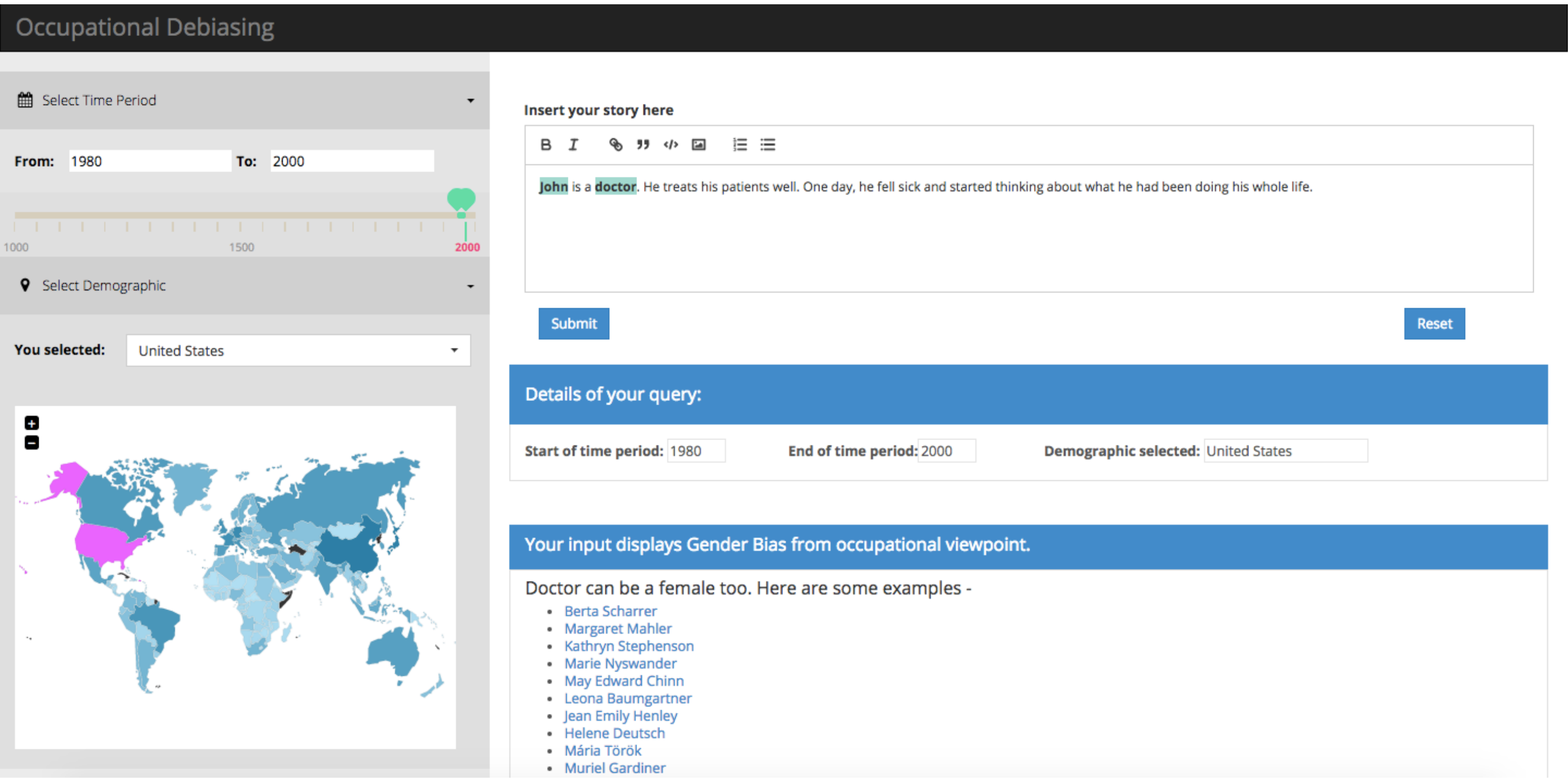}
  \caption{Scenario 1}
  \label{fig:sc1}
\end{figure}

The screen-shots of the interface are represented in \ref{fig:sc1}

\subsection{Scenario 2 : Year 1700-1800 in US}
The story-writer plans to write a story based in United States between the timeframe 1700-1800. He/She uses the story \ref{lst:in-scene-1} and feeds it to the tool.

\begin{figure}[t]
  \centering
  \includegraphics[width=\linewidth]{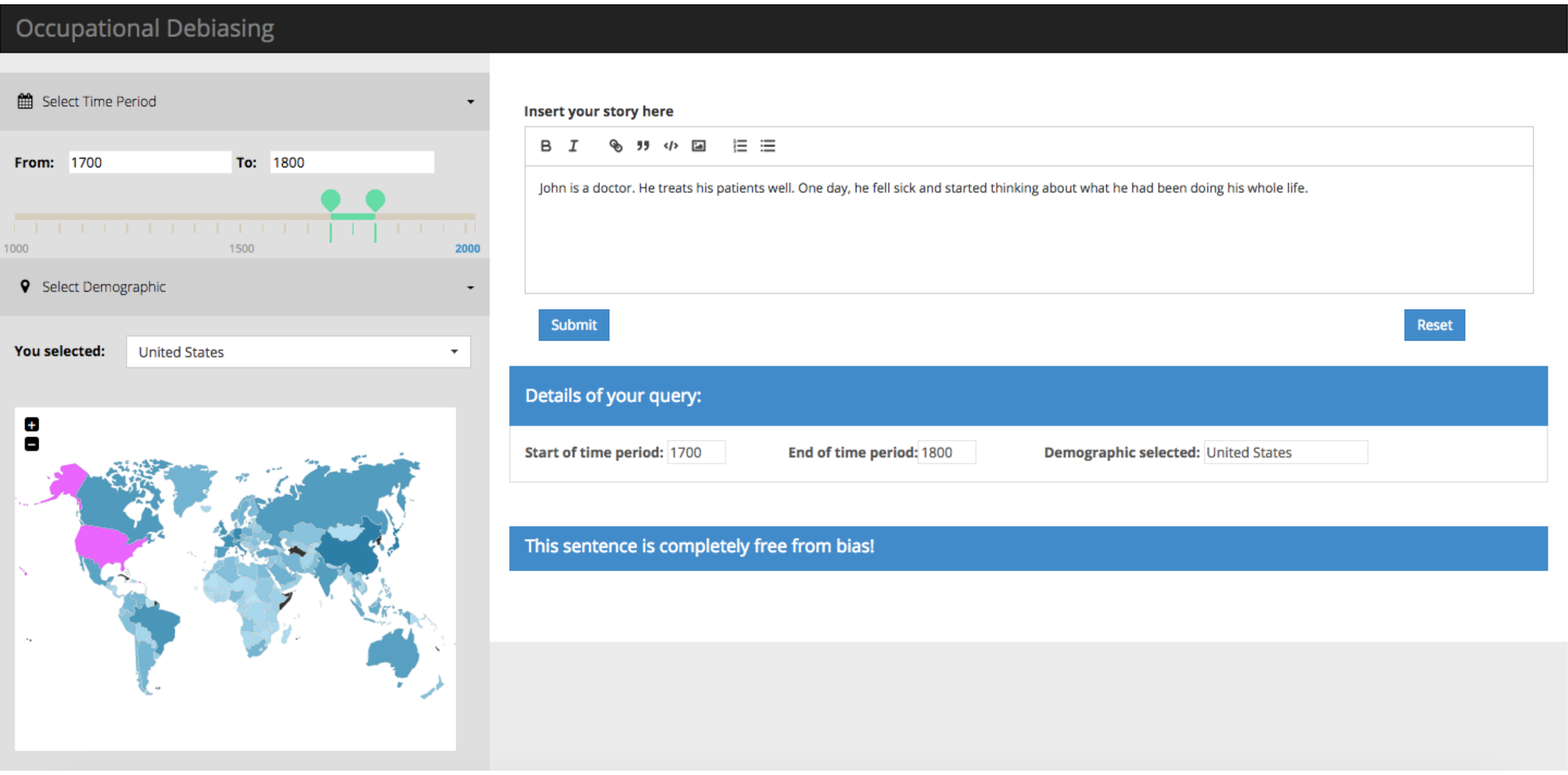}
  \caption{Scenario 2}
  \label{fig:sc2}
\end{figure}

The tool displays no evidences and shows that the story free from bias with occupation point of view. The screen-shot of the interface is shown in \ref{fig:sc2}

\begin{figure}[t]
  \centering
  \includegraphics[width=\linewidth]{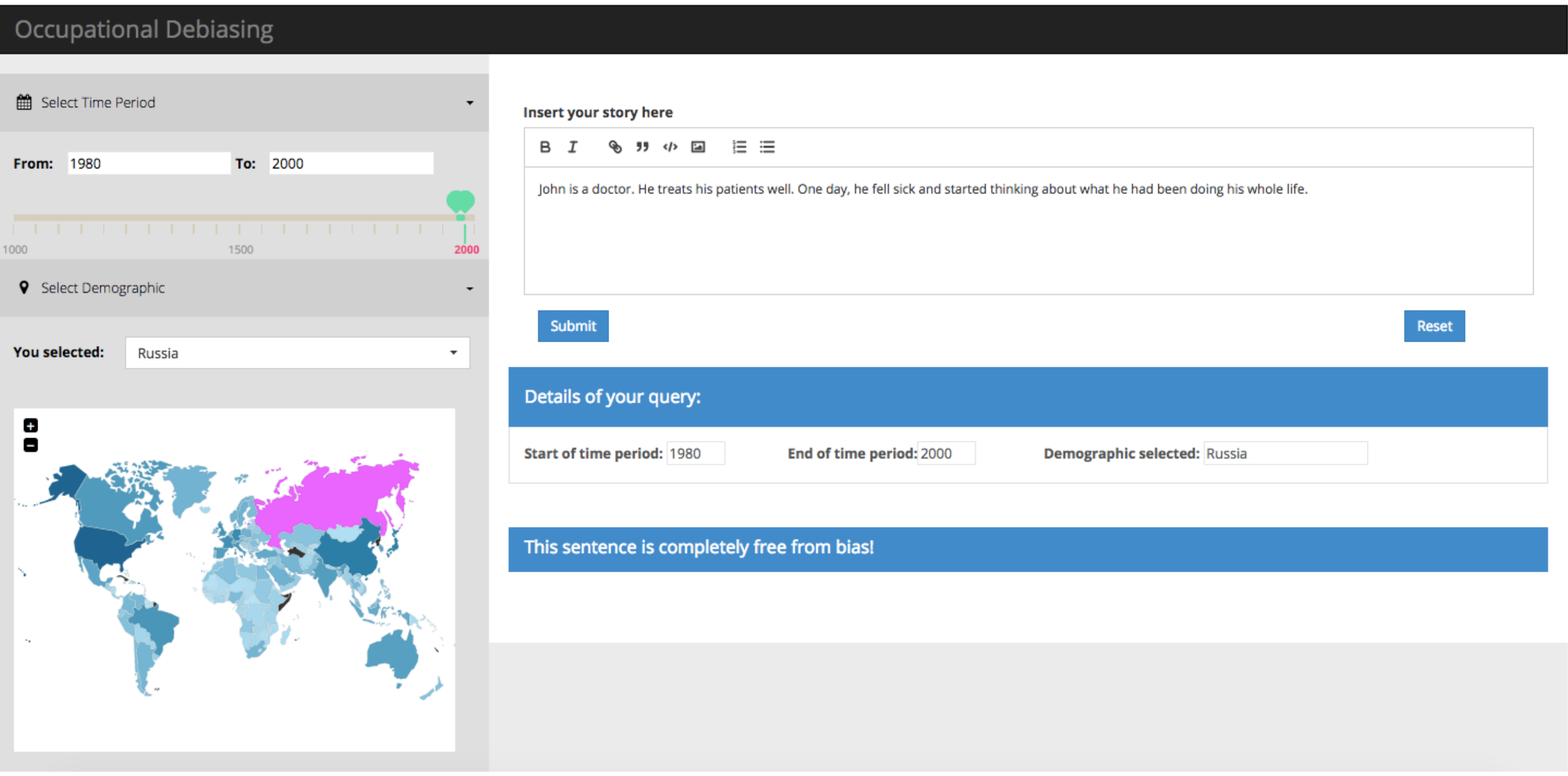}
  \caption{Scenario 3}
  \label{fig:sc3}
\end{figure}

\subsection{Scenario 3 : Year 1980-2000 in Russia}
The story-writer plans to write a story based in Russia between the timeframe 1980-2000. He/She uses the story \ref{lst:in-scene-1} and feeds it to the tool.

The tool displays no evidences and shows the story free from bias with occupation point of view. The screen-shot of the interface is shown in \ref{fig:sc3}

\emph{Hence, the observation is that when we change the year and location parameters in the tool, the tool can automatically respond to the change. Therefore the system is sensitive to the subjectivity of bias in various cultural contexts and timeframes.}

\section{Discussion}

The goal of our system is to be able to remove occupational hierarchy articulated in textual stories.  It is common in movies, novels \& pictorial depictions to show man as boss, doctor, pilot and women as  secretary, nurse and stewardess. In this work, we presented a tool which detects occupations and understand hierarchy and then generate pieces of evidences to show that counter-factual evidences exist. For example, while interchanging (\{male, doctor\}, \{female, nurse\}) to (\{male, nurse\}, \{female, doctor\}) makes sense as there might be evidences in the past supporting the claim but interchanging \{male, gangster\} to \{female, gangster\} might not have evidences in the past for most of the locations. 

To further explain it more, given a sentence -

\begin{lstlisting}[frame= single] 
Jane is a dancer 
\end{lstlisting}

The tool is able to identify that dancer is a gender neutral occupation with a lot of counter-factual evidences of males being dancers in the past. It presents these pieces of evidences to the user to be able to fix this sentence in a guided manner. 

As a future work, we are working on building reasoning systems which automatically regenerate an unbiased version of text.

\section{Conclusion}

Occupation De-biasing is a first-of-a-kind tool to identify possibility of gender bias from occupation point of view, and to generate pieces of evidences by responding to different cultural contexts. Our future work would involve exploring other dimensions of biases and have a more sophisticated definition of bias in text.

\bibliography{acl2018}
\bibliographystyle{acl_natbib}

\end{document}